\documentclass[letterpaper, 10 pt, conference]{ieeeconf}  % Comment this line out if you need a4paper

\IEEEoverridecommandlockouts                              % This command is only needed if 
\usepackage{subfiles}
\usepackage{amssymb}
\usepackage{makecell}
\usepackage{amsmath}
\usepackage{graphicx}
\usepackage{multirow}
\usepackage{soul}
\usepackage[final]{changes}
\usepackage{soul}
\usepackage{color}
\usepackage[utf8]{inputenc}
\usepackage{hyperref}
\usepackage{cleveref}

\crefformat{section}{\S#2#1#3} % see manual of cleveref, section 8.2.1
\crefformat{subsection}{\S#2#1#3}
\crefformat{subsubsection}{\S#2#1#3}

\DeclareMathOperator*{\argminA}{arg\,min} % 

\title{\LARGE \bf
\emph{RangedIK}: An Optimization-based Robot Motion Generation Method for Ranged-Goal Tasks 
}

\author{Yeping Wang$^{1}$, Pragathi Praveena$^{1}$, Daniel Rakita$^{2}$ and Michael Gleicher$^{1}$% <-this % stops a space
\thanks{$^{1}$Yeping Wang, Pragathi Praveena, and Michael Gleicher are with the Department of Computer Sciences, University of Wisconsin-Madison, Madison, WI 53706, USA $\qquad\qquad\qquad\qquad\qquad\qquad\qquad$
{\tt\small [yeping|pragathi|gleicher]@cs.wisc.edu}}%
\thanks{$^{2}$Daniel Rakita is with the Department of Computer Science, Yale University,
        New Haven, CT 06520, USA
        {\tt\small daniel.rakita@yale.edu}}%
\thanks{This work was supported by a University of Wisconsin Vilas Associates Award and National Science Foundation award 1830242.}% <-this % stops a space
}

\begin{document}

\maketitle
\thispagestyle{empty}
\pagestyle{empty}

%%%%%%%%%%%%%%%%%%%%%%%%%%%%%%%%%%%%%%%%%%%%%%%%%%%%%%%%%%%%%%%%%%%%%%%%%%%%%%%%
\begin{abstract}
Generating feasible robot motions in real-time requires achieving multiple tasks (\textit{i.e.}, kinematic requirements) simultaneously. These tasks can have a specific goal, a range of equally valid goals, or a range of acceptable goals with a preference toward a specific goal. To satisfy multiple and potentially competing tasks simultaneously, it is important to exploit the flexibility afforded by tasks with a range of goals. In this paper, we propose a real-time motion generation method that accommodates all three categories of tasks within a single, unified framework and leverages the flexibility of tasks with a range of goals to accommodate other tasks. Our method incorporates tasks in a weighted-sum multiple-objective optimization structure and uses barrier methods with novel loss functions to encode the valid range of a task. We demonstrate the effectiveness of our method through a simulation experiment that compares it to state-of-the-art alternative approaches, and by demonstrating it on a physical camera-in-hand robot that shows that our method enables the robot to achieve smooth and feasible camera motions.

\end{abstract}

%%%%%%%%%%%%%%%%%%%%%%%%%%%%%%%%%%%%%%%%%%%%%%%%%%%%%%%%%%%%%%%%%%%%%%%%%%%%%%%%
\section{INTRODUCTION}
\begin{figure}[tb]
    \centering
    \includegraphics[width=3.4in]{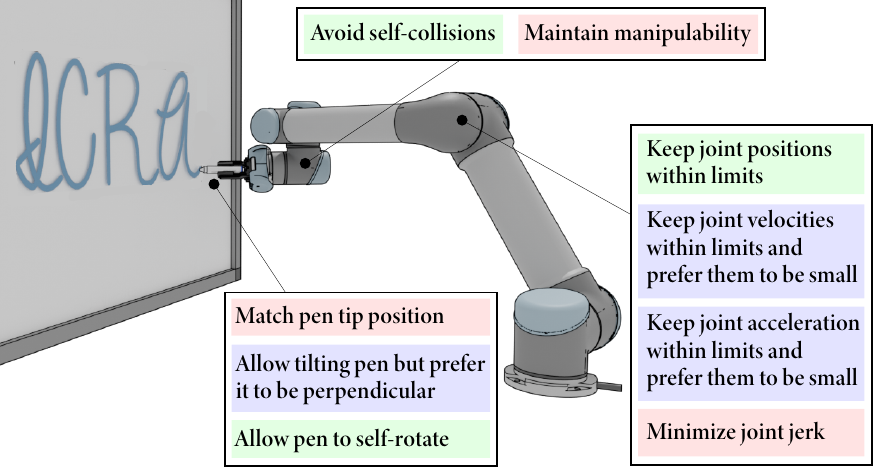}
    \vspace{-6mm}
    \caption{Multiple tasks need to be achieved to generate accurate, smooth, and feasible robot motions in real-time for the whiteboard writing application. These tasks can be classified into three categories according to their flexibility: tasks that have a specific goal (red), tasks that have a range of equally valid goals (green), and tasks that have a range of acceptable goals with a preference toward a specific goal (blue). In this paper, we present a real-time motion synthesis method that can accommodate these three categories of tasks within a single, unified framework.  }
    \label{fig:teaser}
    \vspace{-5mm}
\end{figure}

In real-time robotics applications, the robot needs to calculate how to move at each update to satisfy multiple kinematic requirements simultaneously. As exemplified in Figure \ref{fig:teaser}, a writing application may have several kinematic requirements, which the robot must fulfill to generate accurate and feasible motions. Following Nakamura et al. \cite{nakamura1987task}, we consider the kinematic requirements that the robot must fulfill as \emph{tasks}.
These tasks can be classified into three categories according to their flexibility: (1) A task can have a \emph{specific} goal with limited flexibility, which may cause the robot to lose the capability to achieve other tasks when attempting to accomplish it; (2) A task can have a \emph{range} of equally valid goals, which provides broad flexibility for a robot to accommodate other tasks; and (3) A task can have a \emph{range} of acceptable goals while also showing preference for a \emph{specific} goal, which offers flexibility when needed for more critical tasks while still targeting a specific goal when possible. We refer to tasks in the last two categories as \emph{ranged-goal tasks}, \replaced{in contrast}{as opposed} to the \emph{specific-goal tasks} in the first category.     

In this paper, we introduce a real-time robot motion synthesis method that is able to accommodate specific-goal tasks, ranged-goal tasks with equally valid goals, and ranged-goal tasks with preferred goals within a single, unified framework.  We address the real-time multiple-task motion generation problem through a generalized Inverse Kinematics solver, called \emph{RangedIK}, which incorporates a set of specific-goal or ranged-goal tasks in a weighted-sum non-linear optimization structure.  
\deleted{In particular, our method can make use of flexibility in ranged-goal tasks to improve the accuracy, smoothness, and feasibility of robot motions.} 
\textit{RangedIK} supports flexible ranged-goal tasks in joint space or Cartesian space by utilizing barrier methods with novel loss functions to encode the valid range of a task in optimization.
\added{The multiple-objective optimization structure enables \textit{RangedIK} to leverage the flexibility in ranged-goal tasks to improve the accuracy, smoothness, and feasibility of robot motions.}  \deleted{This enables our method to support flexible ranged-goal tasks in joint space or Cartesian space.}    

% While numerous optimization frameworks can accommodate specific goals or ranged-goals, \textit{i.e.}, through some subset of objective terms, equality constraints, inequality constraints, or variable bounds, we note that    

% We argue that many applications in robotics could benefit from an approach that can accommodate the three classes of tasks specified above 

% and make use of the flexibility in ranged-goal tasks. 

% Specifically, ranged-goal tasks can have a range of \emph{equally valid} goals or a range of \emph{acceptable} goals with a \emph{preferred} goal in it.

% Moreover, optimizing individual poses, \emph{RangedIK} enables a robot to react and adapt to a set of dynamic and complex tasks. 

This paper builds on prior work \cite{rakita2018relaxedik, rakita2020analysis}, which provides a per-instant pose optimization method known as \emph{RelaxedIK} that optimizes single poses to achieve certain accuracy objectives without sacrificing motion feasibility. 
This paper extends the \emph{RelaxedIK} method by leveraging the flexibility \replaced{offered by}{in} ranged-goal tasks.  
The contributions of this paper are threefold: 
(1) a method to incorporate ranged-goal tasks in a multiple-objective optimization-based motion generation structure \added{using novel parametric loss functions} (\cref{sec:shaping} and \cref{sec:parametric});
(2) a set of specific-goal or ranged-goal task functions to generate accurate, smooth, and feasible robot motions (\cref{sec:task_functions}); and
(3) empirical evidence that ranged-goal tasks can create flexibility for improved accomplishment of other tasks (\cref{sec:evaluation}). We provide an open-source implementation of our proposed method\footnote{\url{https://github.com/uwgraphics/relaxed\_ik\_core/tree/ranged-ik}}.

To assess the efficacy of our method, we compare \textit{RangedIK} to \emph{RelaxedIK} \cite{rakita2018relaxedik, rakita2020analysis} and \emph{Trac-IK} \cite{beeson2015trac} on applications \replaced{with Cartesian tolerances, which create ranged-goal tasks to maintain end-effector poses within the tolerances}{with position and rotation tolerances} (\cref{sec:evaluation}). 
Our results suggest that \emph{RangedIK} generates more accurate, smooth and feasible motions than \emph{RelaxedIK}, which does not utilize the flexibility \replaced{offered}{in} by ranged-goal tasks. 
Our evaluation also shows that both \emph{RangedIK} and \emph{Trac-IK} generate valid motions within \added{the specified} tolerance, but our approach produces smoother and more feasible motions than \emph{Trac-IK}. % While \emph{Trac-IK} can only handle Cartesian tolerances, where are ranged-goal tasks in Cartesian space, 
To showcase the generality of our method, we also demonstrate that our method enables a camera-in-hand robot to generate stable videos (\cref{sec:demonstratin}). 
Finally, we conclude this paper with a discussion of the limitations and implications of this work (\cref{sec:discussion}).

% in which keeping the end-effector pose within the tolerances is a ranged-goal task.   

\section{RELATED WORK}
Our work builds upon prior works in semi-constrained motion planning, inverse kinematics, and barrier methods.

\subsection{Semi-Constrained Motion Planning}

Motion planning algorithms are widely used to enable end-effector path following. Semi-constrained motion planners have been presented to generate robot motions in applications \replaced{that have Cartesian tolerances, \emph{i.e.}, where}{that don't require} accurate \added{end-effector} movements in all six degrees of freedom\added{ are not required}. \emph{Descartes} \cite{Descartes, de2017cartesian} is an open-source search-based semi-constrained motion planner released by the ROS-Industrial community. Recently, Malhan et al.  \cite{malhan2022generation} presented an iterative graph construction method to find trajectories that satisfy constraints in Cartesian or joint space. Besides the graph-based methods mentioned above, sampling-based methods have also been used in semi-constrained motion planners \cite{cefalo2020opportunistic, berenson2011task}.  
% A soft task constraint allows an interval of feasible values while favoring a given exact value \cite{kunz2012manipulation}

While semi-constrained motion planners can effectively generate motions for path following, they are not appropriate in time-sensitive, real-time scenarios such as teleoperation. In this work, we use single pose optimization for real-time motion generation. Prior works have shown that single pose optimization is effective for generating motions in real-time scenarios such as teleoperation \cite{rakita2017motion} and camera control \cite{rakita2018autonomous}. 

\subsection{Inverse Kinematics}
Semi-constrained motion planners often call an Inverse Kinematics (IK) solver \textit{repeatedly} to get the joint positions that produce desired end-effector poses. Some emerging IK solvers can handle Cartesian tolerances and only need to be called once by the semi-constrained motion planners. \emph{Trac-IK} \cite{beeson2015trac} handles Cartesian tolerances by redefining Cartesian errors. Such an approach enables Trac-IK to be incorporated with a motion planner \cite{gonccalves2019grasp}, but results in choppy motions when generating motion in real-time (more explanations are in \cref{sec:comparisons}).
In this paper, we present a method that generates smooth motions not only for applications with Cartesian tolerances, but also other ranged-goal tasks.

In our work, we use the term \emph{``task''} from task-priority IK \cite{nakamura1987task, chiacchio1991closed, chiaverini1997singularity, mansard2009versatile}, which utilizes joint redundancy to handle a stack of tasks (kinematic requirements). The task priority framework has been extended to incorporate inequality tasks \cite{kanoun2011kinematic, escande2010fast}. 
Task-priority IK solvers can handle multiple tasks simultaneously by projecting a lower-priority task into the null-space of a higher-priority task, but this approach requires the robot to have sufficient kinematic redundancy for the null-space projections. Our method, on the other hand, can manage \replaced{multiple}{many,} and potentially competing\deleted{,} tasks without the need for such redundancy requirements.

Task-priority IK \replaced{approaches utilize}{has} a strict task hierarchy, which can sometimes be too conservative \cite{salini2011synthesis} or challenging to establish \cite{liu2016generalized}. Meanwhile, non-strict task hierarchies are more flexible and are usually handled by weighted-sum strategies in optimization. Rakita et al. \cite{rakita2018relaxedik} demonstrated the benefits of using multiple-objective optimization to combine motion accuracy tasks with feasibility tasks such as self-collision avoidance. Building on this prior work, our approach leverages the flexibility of ranged-goal tasks to generate accurate, smooth, and feasible motions.  

% Non-strict hierarchy \cite{liu2016generalized} 
% minimizes a weighted sum of objectives subject \cite{BOUYARMANE2011USING, salini2011synthesis}

\subsection{Barrier Methods}

In non-linear optimization, barrier methods convert an inequality constraint to a positively-weighted ``barrier" in the loss function to prevent solutions from leaving feasible regions \cite{forsgren2002interior}. Barrier functions have been widely used in robotics for real-time optimization-based controllers \cite{ames2016control, chen2020guaranteed}, model predictive controllers \cite{grandia2019feedback, feller2016relaxed} and motion planners \cite{yang2019sampling, saveriano2019learning}.
In this work, we apply barrier functions to incorporate ranged-goal tasks in a multiple-objective optimization-based IK structure. 

A strict definition of barrier functions requires them to go to infinity when a solution is close to an inequality constraint boundary. However, limitations of such ``strict" barrier functions have been noted in prior work \cite{grandia2019feedback, feller2016relaxed}; strict barrier functions are only defined over the feasible space and their derivatives go to infinity as one approaches the constraint boundary. 
Therefore, Feller and Ebenbauer \cite{feller2016relaxed} proposed a relaxed logarithmic barrier function that has a suitable penalty term outside of the feasible space to overcome the downsides of strict barrier functions. While this function was designed to possess desired properties for model predictive control, in this paper, we \added{design and} utilize relaxed barrier functions to restrict \replaced{kinematic}{ranged-goal} task values in valid ranges\added{ for real-time motion generation}.

\section{TECHNICAL OVERVIEW}
In this section, we provide notation for our problem and an overview of the optimization structure in our method. 

\subsection{Problem Formulation}

Consider an $n$ degree of freedom robot whose configuration is denoted by $\mathbf{q} \in \mathbb{R}^n$. A \emph{task} (kinematic requirement) is described using $\chi(\mathbf{q}) \in \mathbb{R}$. The tasks can be classified in three categories according to the type of their goals. 

A \emph{Specific-Goal Task} has a \replaced{single valid}{specific} goal, \emph{e.g.}, a pose matching task that requires a robot's end-effector to match a given goal pose.
For such tasks, the value of the task function $\chi(\mathbf{q})$ should match the goal value $g(t)$ at time $t$:
\begin{equation}
    \chi(\mathbf{q}) = g(t)
\end{equation} 

A \emph{Ranged-Goal Task with Equally Valid Goals} allows for multiple valid goals within an interval. For instance, all joint positions are equally valid as long as they are within the joint limits of a robot. 
\deleted{Some manipulation tasks have Cartesian tolerances, \textit{i.e.}, an end-effector pose is valid as long as it's within the tolerances. }
For such tasks, the value of the task function should fall within an interval defined by the lower bound $l(t)$ and the upper bound $u(t)$:

\begin{equation}
   l(t) \leq \chi(\mathbf{q}) \leq u(t)
\end{equation} 

A \emph{Ranged-Goal Task With a Preferred Goal} has a range of acceptable goals while showing a preference for a specific goal. For instance, the joint acceleration of a robot should be within its motor's limits, but we would prefer the acceleration to be as small as possible to minimize energy usage. This type of task is defined by an acceptable interval $[l(t), u(t)]$ and a preferred goal $g(t)$.
%, where the value of the task function is expected to be close to the preferred goal and fall within the interval:
\begin{equation}
    \left\{
    \begin{aligned}
         & \chi(\mathbf{q}) \approx g(t) \\
         & l(t) \leq \chi(\mathbf{q}) \leq u(t)
    \end{aligned}
    \right.
\end{equation} 

The goal of our work is to compute a joint configuration $\mathbf{q}$ to achieve a set of tasks $\{\chi_1(\mathbf{q}), \chi_2(\mathbf{q}),..., \chi_m(\mathbf{q})\}$ \replaced{as best as possible even when competing tasks arise, such as moving a robot to a goal pose while also minimizing energy usage.}{Sometimes the tasks are in conflict, \textit{e.g.}, the task to move a robot to a goal pose and the task to minimize energy usage are conflicting. Our goal is to generate a solution $\mathbf{q}$ that achieves the set of tasks as best as possible.} 
A task can be defined in joint space (\textit{e.g.}, minimizing joint acceleration), Cartesian space (\textit{e.g.}, matching end-effector poses), or a task-relevant space (\textit{e.g.}, keeping an object in view for a camera-in-hand robot). 
\replaced{In real-time applications, f}{F}uture goals $g(t+\delta)$ or goal ranges $[l(t+\delta), u(t+\delta)]$ are unknown at time $t$, for any $\delta > 0$.  

\subsection{Non-linear Optimization}
We formulate the problem posed above as a constrained non-linear optimization problem:
\begin{equation}
\begin{aligned}
    \mathbf{q}^* & = \argminA_\mathbf{q} F(\boldsymbol{\chi}(\mathbf{q})) 
    & s.t. \,\,\, l_i \leq q_i \leq u_i, \,\,\, \forall i
\end{aligned}
\end{equation}
% \mathbf{c(q)} \leq \mathbf{0}, \,\,\,
% $\mathbf{c}$ is a set of inequality constraints, 
where $l_i$ and $u_i$ are the lower and upper bounds of the $i$-th robot joint, and 
$\boldsymbol{\chi}$ is a set of tasks to be achieved.
$F(\boldsymbol{\chi}) \in \mathbb{R}$ is the weighted sum of the loss function for each task: 
\begin{equation}
\label{eq:weighted_sum}
    F(\boldsymbol{\chi}) = \sum_{j=1}^J w_j f_j(\chi_j(\mathbf{q}))
\end{equation} 

Here, $J \in \mathbb{Z}^+$ is the total number of tasks to be achieved and $w_j \in \mathbb{R}$ is the weight value for the $j$-th task. $\chi(\mathbf{q}) \in \mathbb{R}$ is a task function that has a specific goal value or a goal range. $f(\chi) \in \mathbb{R}$ is a parametric loss function that calculates the error between a task value $\chi$ and the specific goal value or the goal range.  We will describe the parametric loss functions and the task functions that we utilize in \cref{sec:technical_details}.

\section{TECHNICAL DETAILS} \label{sec:technical_details}
In this section, we provide details on how we incorporate ranged-goal tasks in a multiple-objective optimization structure. We first introduce three basic loss functions that are used to normalize or regularize task values. Later, we use the basic loss functions as building blocks to design parametric loss functions $f(\chi)$ for specific or ranged-goal tasks. Finally, we detail task functions $\chi$ that are used to generate accurate, smooth and feasible robot motions.

\subsection{Basic Loss Functions} \label{sec:shaping}

\subsubsection{Gaussian} To combine multiple tasks, it is important to normalize each task such that their values are over a uniform range \cite{rakita2018relaxedik}. One common normalization method is the negative Gaussian function: 
\begin{equation} \label{eq:gaussian}
    % f(\chi, g) = -\exp\left(-\frac{(\chi-g)^2}{2c^2}\right)
    f(\chi, g) = -e^{-(\chi-g)^2/2c^2}
\end{equation}
where $g$ is a goal value and $c$ is the standard deviation, which determines the spread of the Gaussian (Figure \ref{fig:shapes}-A). 
% The Gaussian function provides a smooth gradient that drives solutions to 0.

\subsubsection{Wall} Another normalization method involves uniformly scaling values based on the lower bound $l$ and upper bound $u$: \replaced{$\chi' = (2\chi-l-u)/(u-l)$.}
{\begin{equation} \label{eq:wall_normalization}
     \chi' = \frac{\chi-(l+u)/2}{(u-l)/2}
     % \chi' = (2\chi-l-u)/(u-l)
\end{equation}}
After scaling, $\chi'$ is within the interval $[-1,1]$. We present a \added{continuous} function to impose a large penalty when $\chi'$ is close to or outside the boundaries of the interval and a nearly \replaced{zero}{equally small} penalty when $\chi'$ is within the interval:

\begin{equation}
    \label{eq:wall}
    % f(\chi, l, u) = a_1\left[1-\exp \left( - \frac{\chi^{\prime n}}{b^n} \right) \right] \\
    f(\chi, l, u) = a_1\left(1-e^{ - \chi^{\prime n}/{b}^n} \right) 
\end{equation}  

As shown in Figure \ref{fig:shapes}-B, the function has ``walls" at the boundaries, hence it is named the \emph{Wall} function. In Equation \ref{eq:wall}, $a_1 \in \mathbb{R}$ determines the wall ``height", $b \in \mathbb{R}$ determines the wall ``locations''\footnote{We computed $b$ by solving $ 1 - \exp \left( - 1/b^n \right) = 0.95$ to impose 0.95 of the maximum penalty at boundaries.} and $n \in 2\mathbb{Z}^+$ determines the steepness of the walls.

\subsubsection{Polynomial} Neither the Gaussian nor the Wall function provides sufficient derivatives when solutions are far away from their goal or goal range. \added{The non-sufficient derivatives make optimization sensitive to the initial guess, causing it stuck before reaching the optimal point. }\replaced{On the other hand}{Meanwhile}, polynomials provide \replaced{sufficient}{consistent} gradients \added{everywhere} that point towards a goal $g$: 
\begin{equation} \label{eq:polynomial}
    f(\chi, g) = a_2 (\chi-g)^m
\end{equation}
where $ m \in 2\mathbb{Z^{+}}$ is the degree of the polynomial and $a_2$ determines the severity of the penalty when a value is away from the optimal point (Figure \ref{fig:shapes}-C). 

\subsection{Parametric Loss Functions} \label{sec:parametric}
\replaced{Below}{In this section}, we use the basic loss functions in \cref{sec:shaping} as building blocks to design a loss function for each \replaced{task category}{category of tasks}. \added{All of these parametric loss functions are continuous and smooth}.
\subsubsection{Specific-goal tasks}
To enable task values to match a specific goal $g$, Rakita et al. \cite{rakita2018relaxedik, rakita2020analysis} combine the Gaussian and polynomial functions to design the \emph{Groove} loss function:
\begin{equation}
%\small
 % f_{g} (\chi, g, \Omega) = -\exp\left(-\frac{(\chi-g)^2}{2c^2}\right) + a_2 (\chi-g)^m
 f_{g} (\chi, g, \Omega) = -e ^ {-(\chi-g)^2/2c^2} + a_2 (\chi-g)^m
\end{equation}
Here, the scalar values $c$, $a_2$, and $m$ form the set of parameters $\Omega$. As shown in Figure \ref{fig:shapes}-D, the \emph{Groove} function has a narrow ``groove'' around the goal to steer values towards the goal and a more gradual descent away from the goal for better integration \replaced{with}{of} other objectives. 
%The transitions between the Gaussian and polynomial regions are continuous and smooth. 

\subsubsection{Ranged-goal tasks with equally valid goals}
Tasks with a continuous range $[l, u]$ of equally valid goals require a loss function that imposes nearly zero penalties if task values are within the range and large penalties if task values are close to or outside the boundaries. \replaced{Moreover, the function should have sufficient gradients outside the boundaries. To satisfy these requirements,}{The Wall function presented in \cref{sec:shaping} satisfies these two requirements. However, it imposes similarly large penalties for every task value outside of the boundary, causing a solution to be stuck if it is outside the boundary. To address this problem,} we present the \emph{Swamp} loss function, which is a Wall function surrounded by a polynomial: 
\begin{equation}
% \small
   % f_{r} (\chi, l, u, \Omega) =  (a_1 + a_2 \chi^{\prime m})  \left[1-\exp \left( -\frac{\chi^{\prime n}}{b^n} \right) \right] - 1  
   f_{r} (\chi, l, u, \Omega) =  (a_1 + a_2 \chi^{\prime m})  \left(1-e^{-\chi^{\prime n}/b^n} \right) - 1  
\end{equation}
%\begin{align}
%   & \chi' = \frac{\chi-(l+u)/2}{(u-l)/2} \\
%   & 1 - \exp \left( - \frac{1}{b^n} \right) = p
% \end{align}
Here, the scalar values $a_1$, $a_2$, $m$, $n$, and $b$ form the set of parameters $\Omega$. $\chi'$ is the scaled task value described in \cref{sec:shaping}. As shown in Figure \ref{fig:shapes}-E, the \emph{Swamp} function has a flat ``swamp'' region to ensure that any solution within the range is equally good. There are steep ``walls" near the boundaries to restrict solutions \deleted{to be }within the range. \deleted{The height of the walls can be adjusted by the parameter $a_1$.} Outside of the walls, the gradual ``funnel'' provides gradients to drive solutions towards the swamp region. 

\subsubsection{Ranged-goal tasks with a preferred goal}
\begin{figure}[!tb]
    \centering
    \includegraphics[width=3.45in]{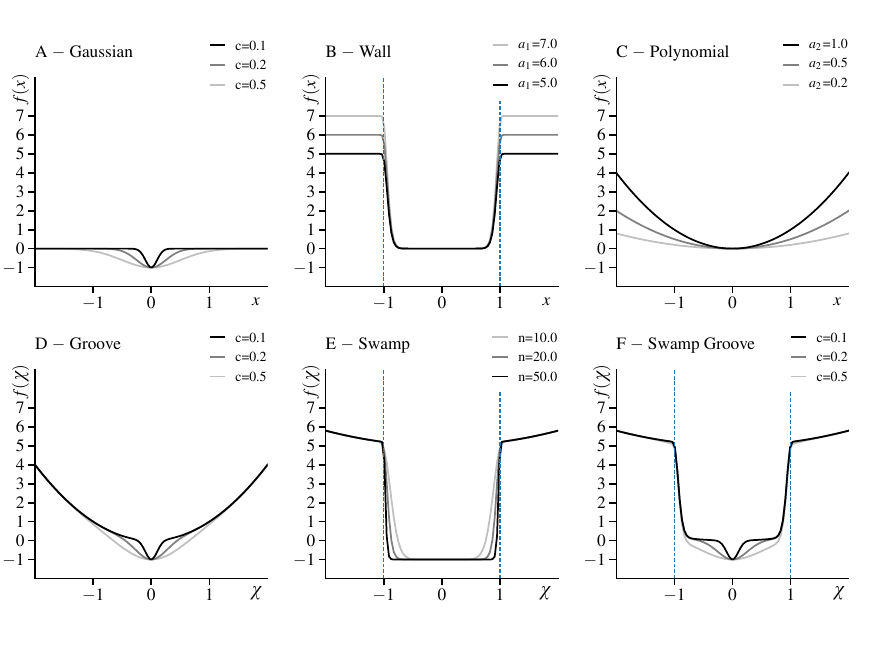}
    \vspace{-10mm}
    \caption{We use the basic loss functions (A, B, C) as building blocks to design parametric loss functions for specific-goal tasks (D), ranged-goal tasks with equally valid goals (E), and ranged-goal tasks with a preferred goal (F). }
    \label{fig:shapes}
    \vspace{-7mm}
\end{figure}

We combine the Gaussian, Wall, and polynomial functions to design the \emph{Swamp Groove} loss function that meets the requirements of tasks with acceptable goals within a range $[l, u]$ and a preferred goal $g$:
\begin{equation}
    \begin{aligned}
   % f_{rg} (\chi, l, u, g, \Omega) = & -\exp{\left(-\frac{(\chi-g)^2}{2c^2}\right)} + a_2 (\chi-g)^m \\
   % & + a_1 \left[1 - \exp \left(  -\frac{\chi^{\prime n}}{b^n} \right)\right]
   f_{rg} (\chi, l, u, g, \Omega) = & -e^{-(\chi-g)^2/2c^2} + a_2 (\chi-g)^m \\
   & + a_1 \left(1 - e^{-\chi^{\prime n}/b^n} \right)
    \end{aligned}
\end{equation}
Here, the scalar values $c$, $a_1$, $a_2$, $m$, $n$, and $b$ form the set of parameters $\Omega$. As shown in Figure \ref{fig:shapes}-F, the \emph{Swamp Groove} function has a ``groove" shape near the preferred goal, steep ``walls" at the boundaries, and a gradual ``funnel" shape outside the walls. We note that the \emph{Swamp Groove} function does not require the preferred goal $g$ to be at the center of the range $[l, u]$ but $g$ must be within the range. 

\subsection{Task Functions} \label{sec:task_functions}
In this section, we describe the mathematical details of task functions used in our work to generate accurate, smooth, and feasible motions. Besides these task functions, additional functions can be introduced for more specialized tasks. 
% e.g., Rakita et al. \cite{rakita2018autonomous} presents objective functions for feasible camera motions. 
Task functions $\chi$ are composed with parametric loss functions $f(\chi)$ described in \cref{sec:parametric} and then combined using Equation \ref{eq:weighted_sum} to form the objective function $F(\boldsymbol{\chi})$. \added{The parameters $\Omega$ of the parametric loss function can be found in our open-source implementation. We estimated the parameters based on visualizations (\textit{e.g.}, Figure \ref{fig:shapes}) and fine-tuned them according to the robot behaviors.} \replaced{The following task functions were adapted from the objective terms in prior work \cite{rakita2018relaxedik}}{We adapt the following task functions from objective terms in prior work \cite{rakita2018relaxedik}}.

\subsubsection{End-effector Pose Matching}
$\chi_P(\mathbf{q}, t, i)$ and $\chi_R(\mathbf{q}, t, i)$ indicate the end-effector's position and rotation error in the $i$-th \replaced{DoF}{dimension}. \replaced{We describe $\chi_P$ and $\chi_R$}{Specifically, both \deleted{the} position and rotation errors are described} with respect to the goal end-effector pose $[\mathbf{p}(t), \mathbf{R}(t)]$. 
\begin{equation}
    \chi_P(\mathbf{q}, t, i) =   \left\{ \mathbf{R}^{-1}(t) \left[ \mathbf{\Psi}_p(\mathbf{q}) - \mathbf{p}(t) \right] \right\}_i
\end{equation}
\begin{equation}
    \chi_R(\mathbf{q}, t, i) =  \left\{ \textbf{S}\left[\mathbf{R}^{-1}(t)  \mathbf{\Psi}_R(\mathbf{q})\right] \right\}_i
\end{equation}
where $\mathbf{\Psi}(\mathbf{q})$ is the robot's forward kinematics function. $\mathbf{\Psi}_p$ and $\mathbf{\Psi}_R$ denote the position and orientation of the robot's end-effector. $\mathbf{S}$ denotes a conversion from a rotation matrix to a vector (scaled-axis) with the direction of the rotation axis and whose norm is the rotation angle. $\{\cdot\}_i$ denotes the $i$-th element of a vector. 
$\chi_P$ and $\chi_R$ can be specific-goal tasks if exact pose matching is desired, but for applications with Cartesian tolerances, \textit{e.g.}, the benchmark applications we evaluate in \cref{sec:evaluation}, $\chi_P$ or $\chi_R$ can be ranged-goal tasks.

\subsubsection{Smooth Motion Generation} We consider the velocity $v$, acceleration $a$, and jerk $j$ of each robot joint separately for the smooth motion generation task. We set limits for each joint as follows ($i$ refers to the $i$-th joint of the robot):  
\begin{equation}
\small
    \begin{aligned}
    \chi_v(\mathbf{q}, t, i) = \dot{\mathbf{q}}_i ;  \,\,\, 
    &\chi_a(\mathbf{q}, t, i) = \ddot{\mathbf{q}}_i ; \,\,\, 
    \chi_j(\mathbf{q}, t, i) = \dddot{\mathbf{q}}_i ;
    \end{aligned}
\end{equation}
For joint velocities and accelerations, we use ranged-goal tasks with preferred goals. Joint velocities or accelerations are \emph{acceptable} if they are within the motor limits and their \emph{preferred} values are zeros. Joint jerks are treated as specific-goal tasks whose goal values are zeros.

\subsubsection{Self-Collision Avoidance} We adapt the collision avoidance method from \cite{rakita2021collisionik} for self-collision avoidance. Each robot link  $\mathbf{l}_i$ is represented as a convex shape (\textit{e.g.}, a capsule). A task function that represents the shortest straight-line distance between the $i$-th and $j$-th link can be written as:
\begin{equation}
    \chi_c(\mathbf{q}, i, j) = \text{dist}(\mathbf{l}_i(\mathbf{q}), \mathbf{l}_j(\mathbf{q}))
\end{equation}
where $\text{dist}()$ is the shortest straight-line distance between two convex shapes and is computed using Support Mapping \cite{kenwright2015generic}. 
To consider self-collisions between all pairs of non-adjacent robot links, a group of ranged-goal tasks are added to the objective function (Equation \ref{eq:weighted_sum}): $\sum_{i=1}^{N-2} \sum_{j=i+2}^{N} f_{r}(\chi_c(\mathbf{q}, i, j),  0.02, \infty,  \Omega)$, where $N$ is the number of robot links. We set 0.02 m as the minimum allowable distance between two non-adjacent robot links. 

\subsubsection{Kinematic Singularity Avoidance} We use the Yoshikawa manipulability measure \cite{yoshikawa1985manipulability} to evaluate how close a configuration $\mathbf{q}$ is to a singularity. The manipulability task is defined as
$\chi_m(\mathbf{q}) = \sqrt{\text{det} \left( \mathbf{J}(\mathbf{q}) \mathbf{J}^T(\mathbf{q}) \right)}$, where $\mathbf{J}$ is the Jacobian matrix that maps joint speeds to end-effector velocities. $\chi_m(\mathbf{q})$ is a specific-goal task and is injected into the loss function $f_{g}(\chi, 1,  \Omega)$ to maximize manipulability.

\section{Evaluation} \label{sec:evaluation}
In this section, we compare \emph{RangedIK} with two alternative approaches, \emph{RelaxedIK} and \emph{TracIK}, to generate motions for applications that allow some tolerances in end-effector poses. These Cartesian tolerances create ranged-goal tasks that maintain end-effector pose within the tolerances.

\subsection{Implementation Details}

Our prototype implementation was based on the open-source \emph{CollisionIK} library\footnote{\url{https://github.com/uwgraphics/relaxed\_ik\_core/tree/collision-ik}}. Our prototype uses the Proximal Averaged Newton-type Method (PANOC)  \cite{stella2017simple} as the optimization solver. All evaluations were performed on an  Intel Core i7-11800H 2.30 GHz CPU with 16 GB RAM.

\subsection{Comparisons} \label{sec:comparisons}

\emph{RelaxedIK} \cite{rakita2018relaxedik} is an optimization-based Inverse Kinematics solver that generates feasible motions given multiple objective terms, such as end-effector pose matching, smooth joint motion, and self-collision avoidance. However, all of the objectives have a single goal. We want to show that by considering ranged-goal tasks, \emph{RangedIK} will generate more accurate, smooth, and feasible motions than \emph{RelaxedIK}. 

Another alternative approach is to generate motions with a widely used Inverse Kinematics solver, \emph{Trac-IK} \cite{beeson2015trac}, which biases the search around the joint space of a given seed value. We provide the seed value as the configuration from the previous update. 
Trac-IK considers Cartesian pose tolerances by mapping pose errors to a discontinuous function:

\begin{equation}
    \label{eq:tracik}
    p_{err}^i = \left\{
    \begin{array}{ll}
    0,     & \text{if \,\,\,\,} l \leq p^i_{err} \leq u\\
    p_{err}^i,      & \text{otherwise}
    \end{array}
    \right.
\end{equation}

Here, $p_{err}^i$ is the pose error in the $i$-th degree of freedom. The discontinuous function makes the robot stop moving when the end-effector pose error is within the interval and leads to a \added{sudden, }large movement once the pose error is outside of the range. Consequently, \replaced{motions generated by Trac-IK can have}{this can cause motions with} large joint accelerations and jerks. 
%However, the official implementation of \emph{Trac-IK} \footnote{https://bitbucket.org/traclabs/trac_ik}  

\subsection{Experimental procedure}

We compared our method to alternative approaches on the four benchmark applications described in Section \ref{sec:benchmark}. The randomly generated path for each benchmark was discretized to 2,000 end-effector goal poses at 30 Hz. The goal poses, along with the Cartesian tolerances, were provided to \emph{RangedIK} and \emph{Trac-IK}, whereas \emph{RelaxedIK} received only goal poses due to its inability to accommodate Cartesian tolerances.
We repeated all benchmarks 10 times using two simulated robots: a 6-DoF Universal Robots UR5 and a 7-DoF Rethink Robotics Sawyer. Our experiment consisted of 480,000 discrete solutions and involved 2 robots, 3 methods, 4 benchmark applications, and 10 randomly generated configurations for each application. 

\begin{table}[tb]
\caption{Cartesian Tolerances}
\label{tab:task_ranges}
\vspace{-4mm}
\begin{center}
\begin{tabular}{lllllll}
\hline
\rule{0pt}{1.05\normalbaselineskip}%
Benchmark & \multicolumn{6}{c}{Tolerances} \\
% & \makecell[l]{$x$ \\ (m)} & \makecell[l]{$y$ \\ (m)} & \makecell[l]{$z$ \\ (m)} & \makecell[l]{rx \\ (rad)} & \makecell[l]{ry \\ (rad)} & \makecell[l]{rz \\ (rad)} \\[0.8mm]
& $x$/m & $y$/m & $z$/m & rx/rad & ry/rad & rz/rad \\[0.5mm]
\hline
\rule{0pt}{1.1\normalbaselineskip}Writing & 0 & 0 & 0 & $\pm\frac{\pi}{6}$ & $\pm\frac{\pi}{6}$ & $\pm\infty$ \\
Spraying & $\pm0.05$ & $\pm0.05$ & 0 & 0 & 0 & 0  \\
Wiping & 0 & 0 & 0 & 0 & 0 & $\pm\infty$  \\
Filling \deleted{Water} & $\pm0.05$ & 0 & $\pm0.05$ & 0 & $\pm\infty$ & 0  \\[0.5mm] 
\hline
\vspace{-10mm}
\end{tabular}
\end{center}
\end{table}

\begin{table*}[!t]
\caption{Results from the Experiment} 
\label{tab:results}
\vspace{-6mm}
\begin{center}
\begin{tabular}{llllllrlcr}
\hline
\rule{0pt}{1.1\normalbaselineskip}%
& Method & \makecell{Mean Pos. \\ Error (m)} & \makecell{Mean Rot. \\ Error (rad)} & \makecell{Mean Joint \\ Vel. (rad/s)} & \makecell{Mean Joint \\ Acc. (rad/s$^2$)} & \makecell{Mean Joint \\Jerk (rad/s$^3$)} & \makecell{Mean Mani-\\pulability} & \makecell{\# Exceed\\Tolerances} & \makecell{Mean Joint \\ Movement (rad)}\\ 
\hline

\rule{0pt}{1.1\normalbaselineskip}%
%%%%%%%%%%%%%%%%%%%% paste data from python %%%%%%%%%%%%%%%%%%%% 

\multirow{3}{*}{\rotatebox[origin=c]{90}{UR5}}%
& \emph{RangedIK} & 0.0021$\pm$0.002 & 0.005$\pm$0.003 & \textbf{0.042$\pm$0.02} & \textbf{0.0269$\pm$0.02} & \textbf{0.224$\pm$0.2} & \textbf{0.0544$\pm$0.01} & 0 & \textbf{16.747$\pm$8.132} \\   
& \emph{RelaxedIK} & 0.0023$\pm$0.002 & 0.007$\pm$0.005 & 0.050$\pm$0.02 & 0.0299$\pm$0.02 & 0.336$\pm$0.3 & 0.0533$\pm$0.01 & 0 & 20.097$\pm$8.796 \\   
& \emph{Trac-IK} & \textbf{1.8e-6$\pm$1.6e-6} & \textbf{1.5e-6$\pm$1.6e-6} & 0.061$\pm$0.04 & 1.9333$\pm$2.48 & 115.536$\pm$149 & 0.0487$\pm$0.01 & 0 & 24.195$\pm$15.918 \\[0.8mm]  
\hline
\rule{0pt}{1.1\normalbaselineskip}%
\multirow{3}{*}{\rotatebox[origin=c]{90}{Sawyer}}%
& \emph{RangedIK} & 0.0014$\pm$0.001 & 0.003$\pm$0.002 & \textbf{0.034$\pm$0.02} & \textbf{0.0265$\pm$0.02} & \textbf{0.290$\pm$0.3} & \textbf{0.1539$\pm$0.04} & 0 & \textbf{15.897$\pm$7.934} \\   
& \emph{RelaxedIK} & 0.0017$\pm$0.001 & 0.006$\pm$0.004 & 0.041$\pm$0.02 & 0.0280$\pm$0.03 & 0.328$\pm$0.4 & 0.1535$\pm$0.04 & 0 & 19.247$\pm$9.783 \\   
& \emph{Trac-IK} & \textbf{1.3e-6$\pm$1.1e-6} & \textbf{2.0e-6$\pm$1.6e-6} & 0.047$\pm$0.03 & 1.4882$\pm$1.79 & 88.930$\pm$107 & 0.1440$\pm$0.05 & 0 & 22.051$\pm$14.029   
\\[0.8mm] 

%%%%%%%%%%%%%%%%%%%%%%%%%%%%%%%%%%%%%%%%%%%%%%%%%%%%%%%%%%%% 

\hline
\multicolumn{9}{l}{\rule{0pt}{1\normalbaselineskip}
The range values are standard deviations. The best value among the three methods for each measure is highlighted in bold.}
\vspace{-7mm}
\end{tabular}
\end{center}
\end{table*}

\subsection{Benchmark} \label{sec:benchmark}

We developed four benchmark applications with Cartesian tolerances (Table \ref{tab:task_ranges}) to compare our method against alternative approaches. 
The first three benchmarks involve manipulation tasks on a whiteboard positioned in front of the robot. The facing angle of the whiteboard is uniformly sampled from $[0, \pi/2]$, with 0 and $\pi/2$ representing a vertical and horizontal whiteboard, respectively.

\subsubsection{Writing} The robot uses a marker as the end-effector to draw curves on the whiteboard. The writing trajectory includes 5 randomly generated cubic Bezier curves.
The application requires the marker tip position to be accurate but allows the marker to tilt (rotational tolerances about the marker's $x$ and $y$ axes) or freely rotate about the marker's principal $z$ axis. 

\subsubsection{Spraying} The robot sprays cleaner on the whiteboard. The spraying positions are uniformly sampled to cover the whiteboard. Some position tolerances parallel to the whiteboard plane (the $xy$ plane) are allowed because the cleaner will be wiped by an eraser in the next application.

\subsubsection{Wiping} The robot wipes the whiteboard with a round eraser, moving along a predefined, lawnmower path to cover the entire whiteboard. The round eraser's rotational symmetry allows the robot to rotate freely about the eraser's principal axis (rotational tolerances about the $z$ axis).

\subsubsection{Filling Water} The robot fills a cup with the water from a faucet and places the cup on a tabletop. The positions of the initial empty cup, the faucet, and the final full cup are uniformly sampled from three $0.2$m$\times0.2$m$\times0.2$m domains within the robot's workspace. A cup can be filled as long as the water flow is within the cup, allowing for some horizontal position tolerances in the $xz$ plane. The cup's rotational symmetry allows the robot to rotate freely about the cup's principal $y$ axis.

\subsection{Metrics}

We used 7 metrics to measure the \emph{accuracy}, \emph{smoothness}, \emph{manipulability} and \emph{validity} of robot motions. Motion \emph{accuracy} was measured using mean position error (m) and mean rotation error (rad), which were measured only in the degrees of freedom without tolerances. We used mean joint velocity (rad/s), mean joint acceleration (rad/s$^2$), and mean joint jerk (rad/s$^3$) to assess motion \emph{smoothness}. Motion \emph{manipulability} was measured by mean Yoshikawa manipulability \cite{yoshikawa1985manipulability}, where a higher value indicates better manipulability. Motion \emph{validity} was measured by the total number of solutions that exceeded the tolerances. We also measured mean movements in joint space, where larger joint movements result in increased wear and tear on the robot. 

\subsection{Results}
\replaced{As shown}{Our results are summarized} in Table \ref{tab:results}\added{, both the non-redundant UR5 and the redundant Saywer robot benefitted from \emph{RangedIK}}. Compared to \emph{RelaxedIK}, our method generated more accurate and smoother robot motions with higher manipulability. Specifically, \emph{RangedIK} could reach better Cartesian accuracy with less joint movement compared to \emph{RelaxedIK}, and all solutions generated by \textit{RangedIK} were within the Cartesian tolerances. 
These results indicate that our method \emph{RangedIK} could leverage the flexibility offered by ranged-goal tasks to improve the quality of generated robot motions. 

\replaced{Our results also show that \emph{RangedIK} could satisfy multiple kinematics requirements simultaneously. Compared to \emph{Trac-IK}, which focuses on generating accurate solutions within joint limits, }{Although the mean position and rotation errors were lower for \emph{Trac-IK} than \deleted{for }\emph{RangedIK},} \emph{RangedIK} \replaced{could generate}{generates} smoother motions with larger manipulability. In particular, due to the discontinuity in Equation \ref{eq:tracik}, \emph{Trac-IK} \replaced{generated}{has} choppy motions with large accelerations and jerks\deleted{ in joint space}.

\section{Demonstration} \label{sec:demonstratin}
The experiment in \cref{sec:evaluation} shows that our method can generate feasible motions for applications with Cartesian tolerances. 
In addition, we demonstrate the effectiveness our method on a camera-in-hand robot that tracks a user's hand to capture clear hand gestures or movements. 

\begin{figure}[!tb]
    \centering
    \includegraphics[width=3.4in]{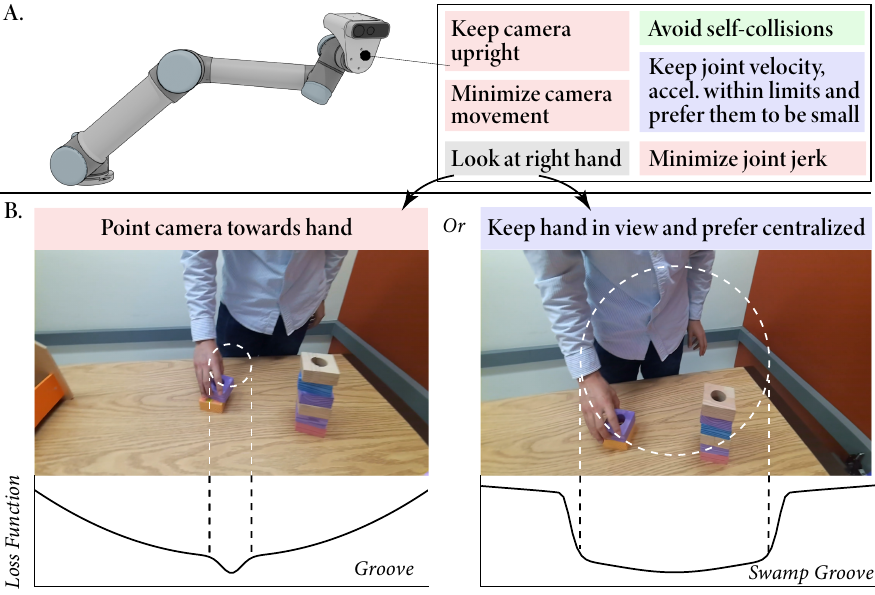}
    \vspace{-5mm}
    \caption{\textbf{A.} We apply our method on a camera-in-hand robot, which requires a set of tasks to enable feasible camera motions. Specific-goal tasks, ranged-goal tasks with equally valid goals, and ranged-goal tasks with a prefer goal are in red, green, and blue, respectively. \textbf{B.} The task to track a user's right hand can be either a specific-goal task (left) or a ranged-goal task with a prefer goal (right). We observed that our method generated smoother videos with flexibility in ranged-goal tasks.}
    \label{fig:camera}
    \vspace{-5mm}
\end{figure}

Prior works \cite{rakita2018autonomous, rakita2019remote} have presented a series of tasks to generate feasible camera motions (Figure \ref{fig:camera}-A). One of these tasks is the ``look-at task" which steers the robot to look at the user's hand. 
While prior works have set a specific goal for this task (Figure \ref{fig:camera}-B left), our method allows us to set a preferred goal and a range of acceptable goals (Figure \ref{fig:camera}-B right). By utilizing the flexibility of the ranged-goal task, our method generates smoother motions to prevent shaky videos. The demonstrations are shown in the supplementary video \footnote{\url{https://github.com/uwgraphics/relaxed\_ik\_core/tree/ranged-ik\#supplementary-video}}.

%\begin{equation}
%    \chi_{\text{lookat}}(t, \mathbf{q}) = \arccos{\frac{|\left(\mathbf{t}(t)-\mathbf{\Psi}_P(\mathbf{q})\right) \cdot \mathbf{v}(\mathbf{q})|}{|\mathbf{t}(t)-\mathbf{\Psi}_P(\mathbf{q})|}}
% \end{equation}

%Here, $\mathbf{t} \in \mathbb{R}^3$ is the target position,  $\mathbf{\Psi}_P(\mathbf{q}) \in \mathbb{R}^3$ is the position of camera, and $\mathbf{v} \in \mathbb{R}^3$ is a unit vector indicating the view direction.

%Our method enables a camera robot to make use of the flexibility of the ranged-goal tasks to generate smoother video.

\section{Discussion} \label{sec:discussion}
In this work, we presented \emph{RangedIK}, a real-time motion generation method that accommodates specific and ranged-goal tasks and leverages the flexibility offered by ranged-goal tasks to generate higher-quality robot motion. Below, we discuss the limitations and implications of this work.

\subsection{Limitations}
Our work has some limitations that highlight directions for future research. 
While our method can generate real-time feasible motions, it can lack foresight because the tasks are defined to find a feasible solution for \emph{now}. 
Extension to this work could investigate ways to incorporate \emph{future} feasibility tasks in our optimization-based structure. 
Also, our method builds on a non-linear optimization formulation that may get stuck in local minima. Future work should explore methods (\textit{e.g.}, warm-start strategies \cite{merkt2018leveraging}) to enable our method to escape local minima.
Finally, \emph{RangedIK} treats all tasks as optimization objectives, which may require scenario-specific tuning of \added{parameters and }weights. Future work can extend our method to include automatic weight-tuning.

% Not a hard constraint, can't guarantee, an outside layer could be added to make sure the output is absolutely valid.

% Setting some ranged-goal is not easy. 

\subsection{Implications}
Our results demonstrate that \emph{RangedIK} can utilize the flexibility in ranged-goal tasks to generate feasible motion on-the-fly.
We believe that such capability makes our method applicable in complex real-life scenarios where many, and potentially competing, tasks need to be achieved simultaneously, such as teleoperation or active camera control. 
Furthermore, the parametric loss functions presented in our work could benefit other optimization-based methods, \emph{e.g.}, offline motion planning or model predictive control.

\bibliography{root}

\begin{thebibliography}{10}
\providecommand{\url}[1]{#1}
\csname url@samestyle\endcsname
\providecommand{\newblock}{\relax}
\providecommand{\bibinfo}[2]{#2}
\providecommand{\BIBentrySTDinterwordspacing}{\spaceskip=0pt\relax}
\providecommand{\BIBentryALTinterwordstretchfactor}{4}
\providecommand{\BIBentryALTinterwordspacing}{\spaceskip=\fontdimen2\font plus
\BIBentryALTinterwordstretchfactor\fontdimen3\font minus
  \fontdimen4\font\relax}
\providecommand{\BIBforeignlanguage}[2]{{%
\expandafter\ifx\csname l@#1\endcsname\relax
\typeout{** WARNING: IEEEtran.bst: No hyphenation pattern has been}%
\typeout{** loaded for the language `#1'. Using the pattern for}%
\typeout{** the default language instead.}%
\else
\language=\csname l@#1\endcsname
\fi
#2}}
\providecommand{\BIBdecl}{\relax}
\BIBdecl

\bibitem{nakamura1987task}
Y.~Nakamura, H.~Hanafusa, and T.~Yoshikawa, ``Task-priority based redundancy
  control of robot manipulators,'' \emph{The International Journal of Robotics
  Research}, vol.~6, no.~2, pp. 3--15, 1987.

\bibitem{rakita2018relaxedik}
D.~Rakita, B.~Mutlu, and M.~Gleicher, ``Relaxedik: Real-time synthesis of
  accurate and feasible robot arm motion.'' in \emph{Robotics: Science and
  Systems}.\hskip 1em plus 0.5em minus 0.4em\relax Pittsburgh, PA, 2018, pp.
  26--30.

\bibitem{rakita2020analysis}
------, ``An analysis of relaxedik: an optimization-based framework for
  generating accurate and feasible robot arm motions,'' \emph{Autonomous
  Robots}, vol.~44, no.~7, pp. 1341--1358, 2020.

\bibitem{beeson2015trac}
P.~Beeson and B.~Ames, ``Trac-ik: An open-source library for improved solving
  of generic inverse kinematics,'' in \emph{2015 IEEE-RAS 15th International
  Conference on Humanoid Robots (Humanoids)}.\hskip 1em plus 0.5em minus
  0.4em\relax IEEE, 2015, pp. 928--935.

\bibitem{Descartes}
\BIBentryALTinterwordspacing
ROS-I. (2015) Descartes—a ros-industrial project for performing path-planning
  on under-defined cartesian trajectories. [Online]. Available:
  \url{http://wiki.ros.org/descartes}
\BIBentrySTDinterwordspacing

\bibitem{de2017cartesian}
J.~De~Maeyer, B.~Moyaers, and E.~Demeester, ``Cartesian path planning for arc
  welding robots: Evaluation of the descartes algorithm,'' in \emph{2017 22nd
  IEEE International conference on emerging technologies and factory automation
  (ETFA)}.\hskip 1em plus 0.5em minus 0.4em\relax IEEE, 2017, pp. 1--8.

\bibitem{malhan2022generation}
R.~K. Malhan, S.~Thakar, A.~M. Kabir, P.~Rajendran, P.~M. Bhatt, and S.~K.
  Gupta, ``Generation of configuration space trajectories over semi-constrained
  cartesian paths for robotic manipulators,'' \emph{IEEE Transactions on
  Automation Science and Engineering}, 2022.

\bibitem{cefalo2020opportunistic}
M.~Cefalo, P.~Ferrari, and G.~Oriolo, ``An opportunistic strategy for motion
  planning in the presence of soft task constraints,'' \emph{IEEE Robotics and
  Automation Letters}, vol.~5, no.~4, pp. 6294--6301, 2020.

\bibitem{berenson2011task}
D.~Berenson, S.~Srinivasa, and J.~Kuffner, ``Task space regions: A framework
  for pose-constrained manipulation planning,'' \emph{The International Journal
  of Robotics Research}, vol.~30, no.~12, pp. 1435--1460, 2011.

\bibitem{rakita2017motion}
D.~Rakita, B.~Mutlu, and M.~Gleicher, ``A motion retargeting method for
  effective mimicry-based teleoperation of robot arms,'' in \emph{Proceedings
  of the 2017 ACM/IEEE International Conference on Human-Robot Interaction},
  2017, pp. 361--370.

\bibitem{rakita2018autonomous}
------, ``An autonomous dynamic camera method for effective remote
  teleoperation,'' in \emph{2018 13th ACM/IEEE International Conference on
  Human-Robot Interaction (HRI)}.\hskip 1em plus 0.5em minus 0.4em\relax IEEE,
  2018, pp. 325--333.

\bibitem{gonccalves2019grasp}
J.~Gon{\c{c}}alves and P.~Lima, ``Grasp planning with incomplete knowledge
  about the object to be grasped,'' in \emph{2019 IEEE International Conference
  on Autonomous Robot Systems and Competitions (ICARSC)}.\hskip 1em plus 0.5em
  minus 0.4em\relax IEEE, 2019, pp. 1--6.

\bibitem{chiacchio1991closed}
P.~Chiacchio, S.~Chiaverini, L.~Sciavicco, and B.~Siciliano, ``Closed-loop
  inverse kinematics schemes for constrained redundant manipulators with task
  space augmentation and task priority strategy,'' \emph{The International
  Journal of Robotics Research}, vol.~10, no.~4, pp. 410--425, 1991.

\bibitem{chiaverini1997singularity}
S.~Chiaverini, ``Singularity-robust task-priority redundancy resolution for
  real-time kinematic control of robot manipulators,'' \emph{IEEE Transactions
  on Robotics and Automation}, vol.~13, no.~3, pp. 398--410, 1997.

\bibitem{mansard2009versatile}
N.~Mansard, O.~Stasse, P.~Evrard, and A.~Kheddar, ``A versatile generalized
  inverted kinematics implementation for collaborative working humanoid robots:
  The stack of tasks,'' in \emph{2009 International conference on advanced
  robotics}.\hskip 1em plus 0.5em minus 0.4em\relax IEEE, 2009, pp. 1--6.

\bibitem{kanoun2011kinematic}
O.~Kanoun, F.~Lamiraux, and P.-B. Wieber, ``Kinematic control of redundant
  manipulators: Generalizing the task-priority framework to inequality task,''
  \emph{IEEE Transactions on Robotics}, vol.~27, no.~4, pp. 785--792, 2011.

\bibitem{escande2010fast}
A.~Escande, N.~Mansard, and P.-B. Wieber, ``Fast resolution of hierarchized
  inverse kinematics with inequality constraints,'' in \emph{2010 IEEE
  International Conference on Robotics and Automation}.\hskip 1em plus 0.5em
  minus 0.4em\relax IEEE, 2010, pp. 3733--3738.

\bibitem{salini2011synthesis}
J.~Salini, V.~Padois, and P.~Bidaud, ``Synthesis of complex humanoid whole-body
  behavior: A focus on sequencing and tasks transitions,'' in \emph{2011 IEEE
  International Conference on Robotics and Automation}.\hskip 1em plus 0.5em
  minus 0.4em\relax IEEE, 2011, pp. 1283--1290.

\bibitem{liu2016generalized}
M.~Liu, Y.~Tan, and V.~Padois, ``Generalized hierarchical control,''
  \emph{Autonomous Robots}, vol.~40, no.~1, pp. 17--31, 2016.

\bibitem{forsgren2002interior}
A.~Forsgren, P.~E. Gill, and M.~H. Wright, ``Interior methods for nonlinear
  optimization,'' \emph{SIAM review}, vol.~44, no.~4, pp. 525--597, 2002.

\bibitem{ames2016control}
A.~D. Ames, X.~Xu, J.~W. Grizzle, and P.~Tabuada, ``Control barrier function
  based quadratic programs for safety critical systems,'' \emph{IEEE
  Transactions on Automatic Control}, vol.~62, no.~8, pp. 3861--3876, 2016.

\bibitem{chen2020guaranteed}
Y.~Chen, A.~Singletary, and A.~D. Ames, ``Guaranteed obstacle avoidance for
  multi-robot operations with limited actuation: A control barrier function
  approach,'' \emph{IEEE Control Systems Letters}, vol.~5, no.~1, pp. 127--132,
  2020.

\bibitem{grandia2019feedback}
R.~Grandia, F.~Farshidian, R.~Ranftl, and M.~Hutter, ``Feedback mpc for
  torque-controlled legged robots,'' in \emph{2019 IEEE/RSJ International
  Conference on Intelligent Robots and Systems (IROS)}.\hskip 1em plus 0.5em
  minus 0.4em\relax IEEE, 2019, pp. 4730--4737.

\bibitem{feller2016relaxed}
C.~Feller and C.~Ebenbauer, ``Relaxed logarithmic barrier function based model
  predictive control of linear systems,'' \emph{IEEE Transactions on Automatic
  Control}, vol.~62, no.~3, pp. 1223--1238, 2016.

\bibitem{yang2019sampling}
G.~Yang, B.~Vang, Z.~Serlin, C.~Belta, and R.~Tron, ``Sampling-based motion
  planning via control barrier functions,'' in \emph{Proceedings of the 2019
  3rd International Conference on Automation, Control and Robots}, 2019, pp.
  22--29.

\bibitem{saveriano2019learning}
M.~Saveriano and D.~Lee, ``Learning barrier functions for constrained motion
  planning with dynamical systems,'' in \emph{2019 IEEE/RSJ International
  Conference on Intelligent Robots and Systems (IROS)}.\hskip 1em plus 0.5em
  minus 0.4em\relax IEEE, 2019, pp. 112--119.

\bibitem{rakita2021collisionik}
D.~Rakita, H.~Shi, B.~Mutlu, and M.~Gleicher, ``Collisionik: A per-instant pose
  optimization method for generating robot motions with environment collision
  avoidance,'' in \emph{2021 IEEE International Conference on Robotics and
  Automation (ICRA)}.\hskip 1em plus 0.5em minus 0.4em\relax IEEE, 2021, pp.
  9995--10\,001.

\bibitem{kenwright2015generic}
B.~Kenwright, ``Generic convex collision detection using support mapping,''
  \emph{Technical report}, 2015.

\bibitem{yoshikawa1985manipulability}
T.~Yoshikawa, ``Manipulability of robotic mechanisms,'' \emph{The international
  journal of Robotics Research}, vol.~4, no.~2, pp. 3--9, 1985.

\bibitem{stella2017simple}
L.~Stella, A.~Themelis, P.~Sopasakis, and P.~Patrinos, ``A simple and efficient
  algorithm for nonlinear model predictive control,'' in \emph{2017 IEEE 56th
  Annual Conference on Decision and Control (CDC)}.\hskip 1em plus 0.5em minus
  0.4em\relax IEEE, 2017, pp. 1939--1944.

\bibitem{rakita2019remote}
D.~Rakita, B.~Mutlu, and M.~Gleicher, ``Remote telemanipulation with adapting
  viewpoints in visually complex environments,'' \emph{Robotics: Science and
  Systems XV}, 2019.

\bibitem{merkt2018leveraging}
W.~Merkt, V.~Ivan, and S.~Vijayakumar, ``Leveraging precomputation with problem
  encoding for warm-starting trajectory optimization in complex environments,''
  in \emph{2018 IEEE/RSJ International Conference on Intelligent Robots and
  Systems (IROS)}.\hskip 1em plus 0.5em minus 0.4em\relax IEEE, 2018, pp.
  5877--5884.

\end{thebibliography}
\bibliographystyle{IEEEtran}

\end{document}